\documentclass[preprint,3p,times,twocolumn]{elsarticle}

\usepackage{amssymb}
\usepackage{multirow}
\usepackage{amsmath}
\usepackage{algorithmic}
\usepackage{float}
\usepackage[ruled,linesnumbered]{algorithm2e}
\usepackage[numbers]{natbib}
\usepackage{color}

\journal{Neurocomputing}

\begin{document}

\begin{frontmatter}



\title{Adaptive and Azimuth-Aware Fusion Network of Multimodal Local Features for 3D Object Detection}

\author[label1,label2]{Yonglin Tian}
\author[label2,label3]{Kunfeng Wang\corref{fn1}}
\author[label4]{Yuang Wang}
\author[label5]{Yulin Tian}
\author[label1]{Zilei Wang}
\author[label2]{Fei-Yue Wang}
\address[label1]{Department of Automation, University of Science and Technology of China, Hefei 230027, China}
\address[label2]{The State Key Laboratory for Management and Control of Complex Systems, \\Institute of Automation, Chinese Academy of Sciences, Beijing 100190, China\\}
\address[label3]{College of Information Science and Technology, Beijing University of Chemical Technology, Beijing 100029, China}
\address[label4]{University of Science and Technology Beijing, Beijing 100083, China}
\address[label5]{North China University of Technology, Beijing 100144, China}
\cortext[fn1]{Corresponding author. Email address: wangkf@mail.buct.edu.cn}


\begin{abstract}
	This paper focuses on the construction of strong local features and the effective fusion of image and LiDAR data for 3D object detection. We adopt different modalities of LiDAR data to generate rich features and present an adaptive and azimuth-aware network to aggregate local features from image, bird's eye view maps and point cloud. Our network mainly consists of three subnetworks: ground plane estimation network, region proposal network and adaptive fusion network. The ground plane estimation network extracts features of point cloud and predicts the parameters of a plane which are used for generating abundant 3D anchors. The region proposal network generates features of image and bird's eye view maps to output region proposals. To integrate heterogeneous image and point cloud features, the adaptive fusion network explicitly adjusts the intensity of multiple local features and achieves the orientation consistency between image and LiDAR data by introducing an azimuth-aware fusion module. Experiments are conducted on KITTI dataset and the results validate the advantages of our aggregation of  multimodal local features and the adaptive fusion network.
\end{abstract}

%


\end{frontmatter}



\section{Introduction}
3D object detection is a fundamental prerequisite for intelligent transportation systems and autonomous vehicles \cite{wang2018intelligent,wang2019social}. Compared with 2D object detection \cite{tian2018training,zhang2019parallel,liu2019towards}, 3D detection needs more spatial information like pose and size. Many studies have achieved precise localization, size and orientation estimation as well as classification of objects in different scenes. Among the state-of-the-art methods for 3D object detection, camera and LiDAR are two of the most widely used sensors. Camera usually has a much longer perceptual distance and higher resolution than LiDAR but loses the depth information, while LiDAR has great superiority in precise measurement of 3D positions. Although some LiDAR-only models \cite{zhou2018voxelnet,yan2018second} have demonstrated good performance on 3D detection datasets like KITTI \cite{geiger2012we}, adopting LiDAR-camera setup is prone to building a more reliable system in practice, especially for far and occluded objects. Many works \cite{chen2017multi,ku2018joint,liang2018deep,qi2018frustum,xu2018pointfusion} have made significant achievements by either using image and LiDAR at different stages or fusing them at pixel and object levels. However, due to the heterogeneous data formats and distinct feature styles, the efficient extraction and effective fusion of LiDAR and image features remain challenging problems. 

Image feature extraction has been well settled by recent works \cite{simonyan2014very,szegedy2015going,he2016deep,lin2017feature,hu2018squeeze} in 2D Convolutional Neural Network (CNN). The key problem of feature extraction in LiDAR-camera setup lies in the processing of LiDAR data. Some works \cite{chen2017multi,ku2018joint,yang2018pixor} project LiDAR point cloud into Front View (FV) or Bird's Eye View (BEV) to obtain image-like data and use 2D CNNs to extract LiDAR features. Some other works \cite{zhou2018voxelnet,yan2018second} voxelize point cloud into regular grids and aggregate the features in each grid. Both two methods achieve efficient point cloud feature generation, but lose information more or less during such quantification.
PointNet \cite{qi2017pointnet} proposes a new way to process 3D data directly, which is promising to maintain richer features in 3D point cloud. Due to the unordered data format, local feature extraction is not as convenient as that in image-like data processing, where relative position can be simply determined by the indexes of pixels. PointNet++ \cite{qi2017pointnet++} addresses this problem by exploiting metric space distances. It promotes the performance on both point cloud classification and segmentation, but is somewhat time-consuming. To make full use of the information in point cloud and achieve efficient feature extraction, we maintain both original point cloud data and projected BEV maps in our network. BEV maps are used to generate proposals with image data. Original point cloud is processed by our ground plane estimation network which generates global-level features for plane regression and pixel-level features for subsequent aggregation. With this hybrid design, we can make coarse but fast proposal generation at the first stage and achieve efficient local feature aggregation for point cloud data based on the 3D Regions of Interest (RoI) at the second stage. Using multiple local features from the image, BEV map and point cloud, proposals can be further processed to get the final detection results.

\begin{figure}
\centering
\includegraphics[scale=0.28]{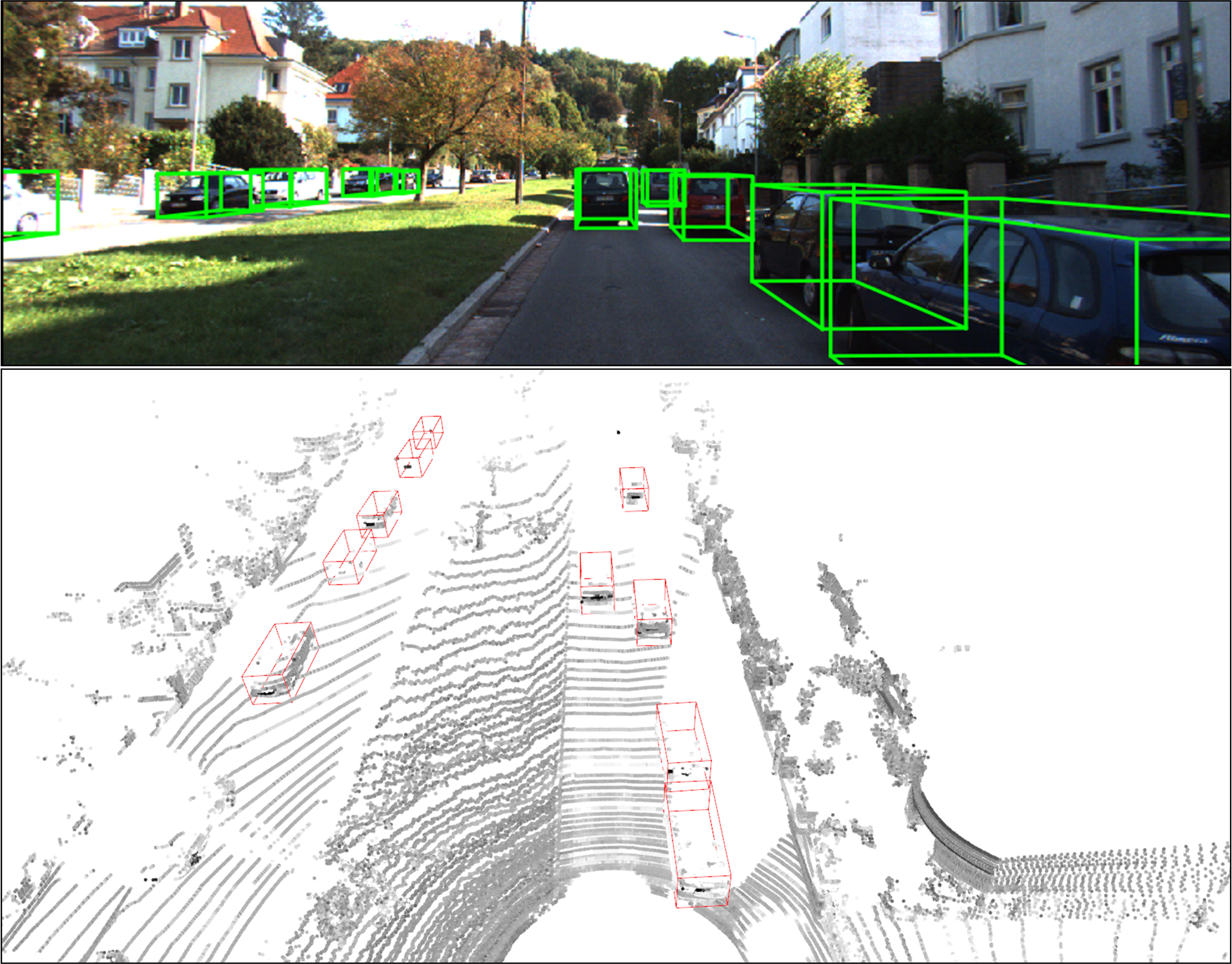}
\caption{3D object detection shown in image (upper one) and LiDAR point cloud (lower one). Our method aggregates local features from image, BEV maps and point cloud which are then fused effectively with our adaptive fusion network.}
\label{3ddet}
\end{figure}

Fusion of multimodal features is another key problem in LiDAR-camera setup systems. Recent works involve pixel-level fusion \cite{liang2018deep,wang2019densefusion} and RoI-based fusion \cite{chen2017multi,ku2018joint,xu2018pointfusion}. Pixel-level fusion matches points between different modalities of data by projecting 3D points to image or BEV plane. It achieves fine-grained fusion style but neglects embedding structural information inside original feature space. The RoI-based fusion method projects 2D \cite{xu2018pointfusion} or 3D \cite{chen2017multi,ku2018joint} RoIs into different spaces and aggregates corresponding parts, which retain both pixel features and local structural information. These fusion methods emphasize more on the correspondence between different features but ignore the difference inside multimodal features. It is a fairly important problem and can be presented in many multimodal systems due to different data formats and feature extractors. For LiDAR-camera setup, image array stores the RGB information of each pixel and is processed by CNNs generally. However, point cloud records the coordinates of each point and needs to be coped with some order-invariant methods. Therefore, the distribution of features from different sources can be significantly varied, so that it is inappropriate to fuse them directly. For example, features with strong magnitude can easily cover weaker ones and dominate the learning of following network. Another neglected problem in RoI-based fusion networks is the inconsistency of orientation of object shown in image and LiDAR data. Due to the effects of camera imaging, the orientation of object shown in image is influenced by the relative position between the object and the camera, while it is only concerned with the object itself in LiDAR data. This kind of mismatch can heavily confuse the orientation estimation. To settle these two problems, we propose an adaptive fusion network that consists of an adaptive weighting network and a spatial fusion module. The adaptive weighting network is inspired by recent works \cite{hu2018squeeze,vaswani2017attention,chen2017sca,woo2018cbam} on feature extraction. We design a feature-wise weighting network to learn the weights for input features in an unsupervised manner, which are multiplied with the corresponding inputs to adjust the distribution of multiple features. It achieves a balance between diverse features and lays the foundation for subsequent fusion. The spatial fusion module tiles 2D feature inputs into 3D space and uses azimuth information to achieve the orientation consistency of image and LiDAR features. The final 3D detection result is illustrated in Figure \ref{3ddet}.

The main contributions of this work can be summarized as follows:
\begin{itemize}
\item  We present a data-driven plane estimation network to achieve fast and precise ground plane prediction.
\item We design a hybrid network that comprises different kinds of feature extractors for both image-like data (image and BEV maps) and point cloud to take full advantage of mature 2D CNNs and the spatial information inside 3D point cloud.
\item We build an adaptive weighting network to dynamically adjust the distribution of different features. It balances the importance of various features for each proposal and promotes the building of a stable detection system.
\item We propose an azimuth-aware fusion module to match the orientation information embedded in image and LiDAR data. It tiles the features into 3D space to approximate the spatial structure of objects and manipulate them there to achieve orientation consistency.
\end{itemize}

All these contributions help to build an effective framework for 3D object detection. Experimental results verify the advantage of our method to utilize point cloud data and the fusion style. 

\begin{figure*}
\centering
\includegraphics[scale=0.5]{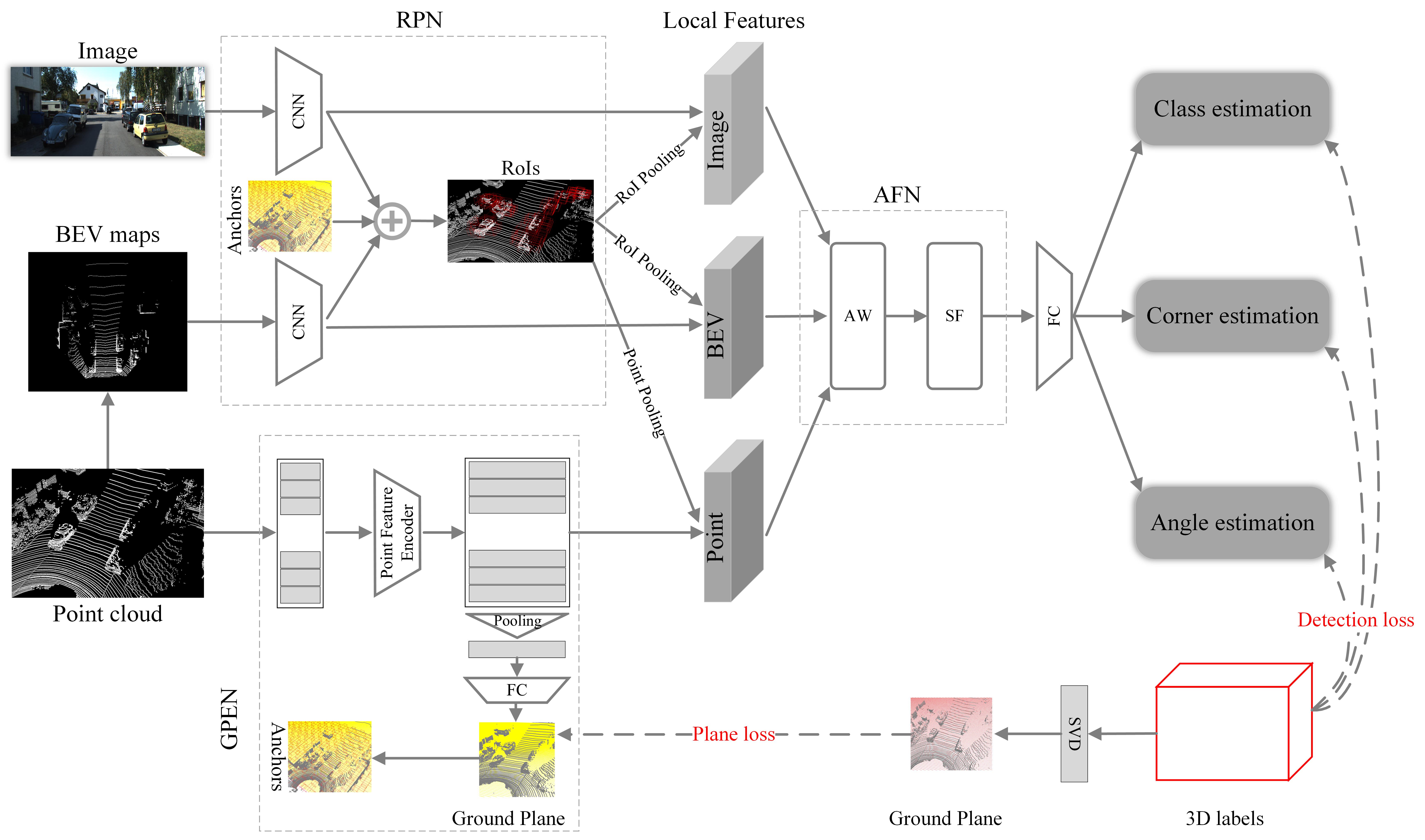}
\caption{The proposed framework for 3D object detection. Main components of our network are Ground Plane Estimation Network, Region Proposal Network and Adaptive Fusion Network. LiDAR point cloud data is firstly processed to predict a ground plane which is used to generate 3D anchors. Lots of region proposals are produced in RPN with both image and projected BEV maps. Different local features from image, BEV maps and point cloud are aggregated in Adaptive Fusion Network. The Adaptive Weighting module learns to adjust the magnitudes of these local features and the Spatial Fusion module extends 2D features to 3D space and aligns the orientation indicated by image and BEV features.}
\label{framework}
\end{figure*}

\section{Related Works}

Image and LiDAR data are two of the most widely used sources for 3D object detection. Various setups of input data and different fusion manners have been explored in previous works.   

\subsection{Image-based approach}
Although the performance of image-based method lags that of LiDAR-involved one, it continues attracting attention due to the economical deployment in autonomous vehicles. Image-based methods can be summarized as monocular and stereo approaches. Monocular approach usually makes 2D predictions of objects on the image plane firstly and relies highly on the geometry priors \cite{chen2016monocular,mousavian20173d} or 3D object models \cite{kundu20183d,chabot2017deep} to make up for the loss of depth information. Inferring 3D location and size from a sigle 2D image is actually an ill-suited problem; however, more satisfactory results can be got with more 2D information introduced to constrain the projection form 3D plane to 2D image, such as bounding boxes \cite{mousavian20173d}, segmentation results \cite{kundu20183d}, depth maps \cite{kundu20183d} and key points \cite{chabot2017deep}. Recently, Thomas et al. \cite{roddick2018orthographic} propose Orthographic Feature Transform (OFT) to transform the features from image space to BEV space and highlight the superiority of the latter in 3D object detection. Using two different cameras simultaneously, stereo approach can predict a more accurate depth map \cite{chen20183d,pham2017robust}. Either concatenating depth map with image features \cite{pham2017robust} or turning it into pseudo LiDAR \cite{chen20183d,wang2019pseudo} has shown better results than monocular approach. Stereo R-CNN \cite{li2019stereo} takes 3D object detection as a learning-aided problem and solves it with the discrepancy of bound boxes and key points between left and right images. Image-based approach has great advantage in the cost and deployment in reality; however, no competitive results have been achieved compared with the LiDAR-involved method.

\subsection{LiDAR-based approach}
LiDAR is one of the sensors used to generate a collection of sparse points in 3D space. To deal with the unordered LiDAR point cloud, many methods reorganize it into regular format. Some of them divide the point cloud into equally spaced 3D voxels and aggregate the points in each voxel into one feature vector. 3D CNN \cite{zhou2018voxelnet,yan2018second,maturana2015voxnet,wu20153d} can be applied to further process these features and generate the final detection. However, this approach is not efficient enough. Most of the voxels are empty due to the sparsity of point cloud. Yan et al. \cite{yan2018second} present a new sparse convolutional middle extractor to speed up the computation and reduce memory usage. Some other methods \cite{qi2016volumetric,su2015multi} apply the 2D CNN to deal with LiDAR data by projecting it into image-like data from different perspectives. This takes the full advantage of existing 2D networks but destroys the structure inside 3D data. PointNet \cite{qi2017pointnet} and PointNet++ \cite{qi2017pointnet++} pave a new way for the processing of point cloud which propose an effective method to handle unordered point set. They directly extract point-level features and aggregate them with a max-pooling to cope with the disordering.

\subsection{Image-LiDAR approach}
Simultaneously taking image and LiDAR data as the source of detection system is prone to giving more accurate performance. One popular way to use multi-sensor data deploys different signals at different stages \cite{qi2018frustum,xu2018pointfusion,shin2018roarnet}. Image is usually used for proposal generation and LiDAR data is used for box refinement. This method often leads to a poor recall because depth is unable to be inferred from image alone and it is easy to miss objects during proposal generation. Another approach \cite{chen2017multi,ku2018joint} projects the LiDAR data into 2D plane, so both image and projected LiDAR data can be used in either proposal generation or refinement network. It improves the recall rate of the proposals but the stereo structure suffers from the projection. In our network, both image and LiDAR data are used in both stages but in different modalities to alleviate the information loss and leverage the mature 2D CNNs. Most of the image-LiDAR approaches fuse these features at  point-level \cite{liang2018deep,wang2019densefusion} or region-level \cite{chen2017multi,ku2018joint,xu2018pointfusion}. Point-level fusion finds the correspondence of points in image and LiDAR data by aid of the 2D-3D projection. It is straight forward but ignores the original structure of features in different spaces. Region-level fusion takes 2D or 3D RoI as the bridge between multiple features and keeps the structural information by cropping local regions in different data. \textcolor{black}{Recently, MMF \cite{liang2019multi} exploits the supplementary effects of multi-task network and gets good performance on KITTI dataset.}

\subsection{Ground plane fitting in point cloud}
There are many classic methods proposed for curve and plane fitting, such as least square method and eigenvalue-based methods \cite{wang2001comparison}. These approaches work well for clean data but are unable to get satisfactory results for ground plane fitting in point cloud. They can be easily misled by the irrelevant points in the LiDAR data such as buildings, trees and other objects. RANSAC \cite{fischler1981random} presents a robust method to fit the model to data with significant noise. Different from these methods, we propose a data-driven subnetwork to estimate the parameters of ground plane which can either be trained separately or be plugged into our detection network.

\section{Proposed Method}
In this section, we describe the architecture of our 3D detector as shown in Figure \ref{framework}. It mainly comprises three parts: Ground Plane Estimation Network (GPEN), Region Proposal Network (RPN) and Adaptive Fusion Network (AFN).

\subsection{Ground plane estimation network}
\begin{figure}
\centering
\includegraphics[scale=0.5]{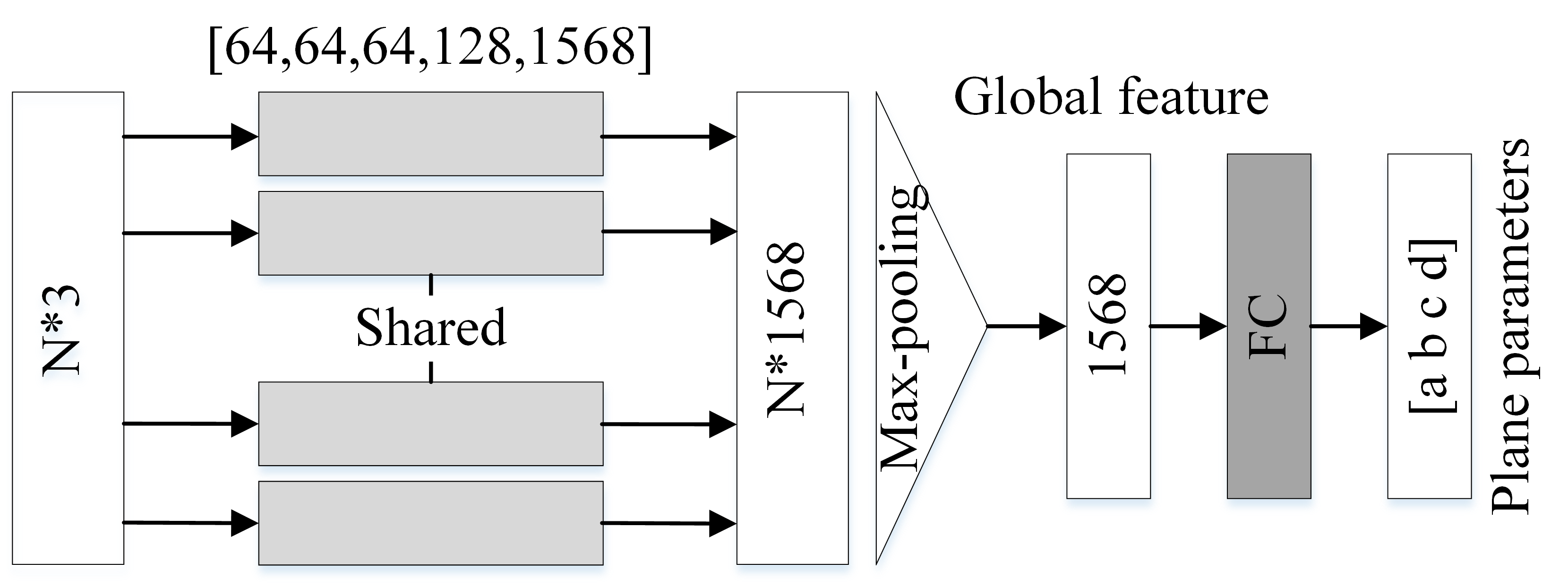}
\caption{The variant of PointNet used to extract features for point cloud.}
\label{PointNet}
\end{figure}

We assume a ground plane here which is reasonable for most cases and adopt a variant of PointNet as our point feature encoder to predict the parameters of the plane. Considering that we only need to detect objects shown in the image, points outside the view of camera are filtered out. We also remove points whose y coordinates are not in range [-1,3] meter to further reduce computation. With points of shape $[N,3]$ as input to our point feature encoder, we transform them into higher dimension and get point features of shape $[N,C]$ which can be used for ground plane parameters estimation or local feature aggregation. The details of our GPEN are shown in Figure \ref{PointNet}. We apply a max-pooling operator to get the global features and use fully-connected layers to regress the plane.

To train GPEN in a supervised manner, we transform the labeled 3D bounding boxes in each frame to parameters of a pseudo ground plane. In our assumption, the ground plane can be formulated as
\begin{align}
\begin{split}
a^{*}x+b^{*}y+c^{*}z=d^{*},
\end{split}
\end{align}%
where $[a^{*},b^{*},c^{*}]$ is the unit normal vector indicating the direction of plane and $d^{*}$ is the distance from the origin to the plane. To get ground plane label, we use Singular Value Decomposition (SVD) to analyze the bounding boxes and produce pseudo ground labels. With a set of $m$ 3D bounding boxes $B=\{b_{1},b_{2},...,b_{m}\}$ in each frame, we can get $4m$ points $P=\{p_{1},p_{2},...,p_{4m}\}$ by gathering the corners on the bottom of each bounding box. All points are then decentered and reformulated as a matrix $\bar{P}$. We use SVD to get the singular vector with the maximum singular value and take it as the unit normal vector of the ground plane. Finally, $d^{*}$ can be determined by putting the center of all points in each frame to Equation (1). The pseudo ground label generation algorithm for a single frame of data is shown in Algorithm 1.

The loss function in our GPEN is composed of two smooth L1 loss :
\begin{align}
\begin{split}
Loss_{gpen}=&\alpha_{v}*smoothL1(Vec,Vec^{*})+\\
&\alpha_{d}*smoothL1(d,d^{*}),
\end{split}
\end{align}%
where $Vec^{*}$ is the ground truth unit normal vector $[a^{*},b^{*},c^{*}]$ and $Vec$ is the predicted vector $[a,b,c]$, $d^{*}$ is the distance from the origin to the ground plane and $d$ is the predicted distance. We use $\alpha_{v}$ and $\alpha_{d}$ to balance these two loss and they are set to 5 and 50 in our experiments.

\begin{algorithm}
\caption{Pseudo ground label generation algorithm}
\label{alg1}
\KwIn{Bounding set: $B=\{b_{1},b_{2},...,b_{m}\}$\;}
\KwOut{Ground plane parameters $[a^{*},b^{*},c^{*},d^{*}]$\;}
\LinesNumbered
Get bottom corner set $P=\{p_{1},p_{2},...,p_{4m}\}$ from bounding box set $B$\;
Calculate the mean of all points: $\bar{p}$=$\frac{1}{4m}\sum_{i=0}^{4m}\left ( p_{i} \right )$\;
Decenter all points with $\bar{p}$: $p_{i} \leftarrow p_{i}-\bar{p}$\;
Refomat all points in a matrix $\bar{P}$\;
Decompose $\bar{P}$ with SVD algorithm: $\bar{P}=U \Sigma V^T$\;
Set the first vector in $U$ as the unit normal vector of the ground plane: $[a^{*},b^{*},c^{*}]=U^T[0]$\;
Calculate $d^{*}$: $d^{*}=[a^{*},b^{*},c^{*}]\cdot[\bar{p}[0],\bar{p}[1],\bar{p}[2]]^T$

\end{algorithm}

\subsection{Region proposal network}
Region Proposal Network (RPN) is first proposed in \cite{ren2015faster}. It is the first stage of our framework, aiming at proposing regions that potentially include objects. Anchors are distributed on the ground plane with an interval of 0.5 meter. The size of anchors is decided by the clustering results on training set and the direction is simply set to 0 and 90 degree with the forward direction.  We assign positive labels to the anchor whose IoU with the groud-truth on BEV plane is greater than 0.5 and negative labels to those whose IoU with ground-truth is lower than 0.3. We take image and BEV maps of point cloud as inputs and predict the binary classification $ RPN_{cls} $ and axis-aligned 3D bounding box estimation $RPN_{reg}$ comprising box center $ RPN_{center} $ and box size $RPN_{size}$ of each proposal. To generate BEV maps, we filter out points that are outside the field of camera and [-0.2, 2.3] meter with relative to the predicted ground plane along the vertical axis.  The horizontal plane is divided into small grids with resolution of 0.1 $\times$ 0.1 $m^{2}$ and a density map is produced by calculating the density of points in each grid. We equally slice the space into five parts along the vertical axis and get five height maps that represent the maximum height of points within each bin. Totally, we get six BEV maps which are used as the substitution for point cloud. 

\textcolor{black}{As shown in Figure 2, we use two convolutional networks with VGG backbone and FPN \cite{lin2017feature} structure to extract the features of image and BEV maps. To fuse these two kinds of features, we adopt anchor-based fusion strategy. We first reduce the number of channels of image and BEV features to 1 with 1 $\times$ 1 convolution. Then, we project each 3D anchor to image plane and BEV plane respectively and crop the corresponding features. These cropped features are resized to the same size of 3 $\times$ 3 and fused by element-wise addition. Finally, we use two branches of fully-connected layers to further process the fused features and generate classification and box regression results for each anchor. After the NMS, we get the proposed regions of interest.} 
Cross-entropy loss and smooth L1 loss are used for classification and box regression respectively. The RPN loss is defined as follows:
\begin{align}
\begin{split}
Loss_{rpn}=&\beta_{c}*cross\_entropy(RPN_{cls},RPN_{cls}^{*})+\\
&\beta_{r}*smoothL1(RPN_{reg},RPN_{reg}^{*}),
\end{split}
\end{align}%
where $RPN_{cls}^{*}$ is the binary label and $RPN_{reg}^{*}$ denotes the ground-truth center and size of the proposal. $\beta_{c}$ and $\beta_{r}$ are set to 1.0 in our experiments.

Taking BEV maps as substitute for original point cloud here is primarily for the fact that proposals are usually far more than objects included in the image or point cloud. A large number of proposals and our following refinement network can well compensate for the loss of useful information introduced by the mapping process from point cloud to BEV maps. So, we emphasize more on efficiency than precision at this stage.

\subsection{Adaptive fusion network}
With the proposals generated by RPN, we propose a novel fusion network to make the final predictions. Both image and LiDAR features are used here. Different from previous fusion-based works that use either BEV maps or point cloud as representation of LiDAR data, we maintain two modalities of LiDAR data at this stage. For clarity, features extracted from BEV maps are denoted as BEV features and those extracted from the original point cloud are denoted as point features in our work. 

\paragraph{Local feature extraction}
Image and BEV features have been generated in RPN using VGG network and FPN. To get features for each proposal, we need to map the 3D boxes to 2D regions on the image plane and BEV plane. This can be formulated as:
\begin{align}
\begin{split}
&Region_{img}=RoIP_{img}(box_{3d}),\\
&Region_{bev}=RoIP_{bev}(box_{3d}),
\end{split}
\end{align}%
where $ROIP_{img}$ ($ROIP_{bev}$) is the RoI pooling operation between the 3D space and image (BEV) space. The corresponding regions in image and BEV features are cropped and then resized into the same size for later fusion. We denote the extracted image and BEV features as $f_{il}$ and $f_{bl}$ respectively which have the shape of [7, 7, 32].

To generate local features in point cloud, we select the points that are located in the range of the proposal from the point cloud and get a tensor with shape $[N_{i},C]$. $N_{i}$ indicates the number of points in the $i$-th proposal and $C$ is the dimension of features generated in GPEN. $N_{i}$ varies considerably due to the uneven distribution of points in point cloud. For convenience of later processing, we randomly sample $M$ points for proposals that hold more than $M$ points, and pad points with zeros for those that possess less than $M$ points. With point pooling, we get a set of point features $f_{p}$ with shape $[M,C]$. And after a max-pooling operator, the point features are aggregated into local features $f_{pl}$ with respect to each proposal.
\begin{figure}
\centering
\includegraphics[scale=0.40]{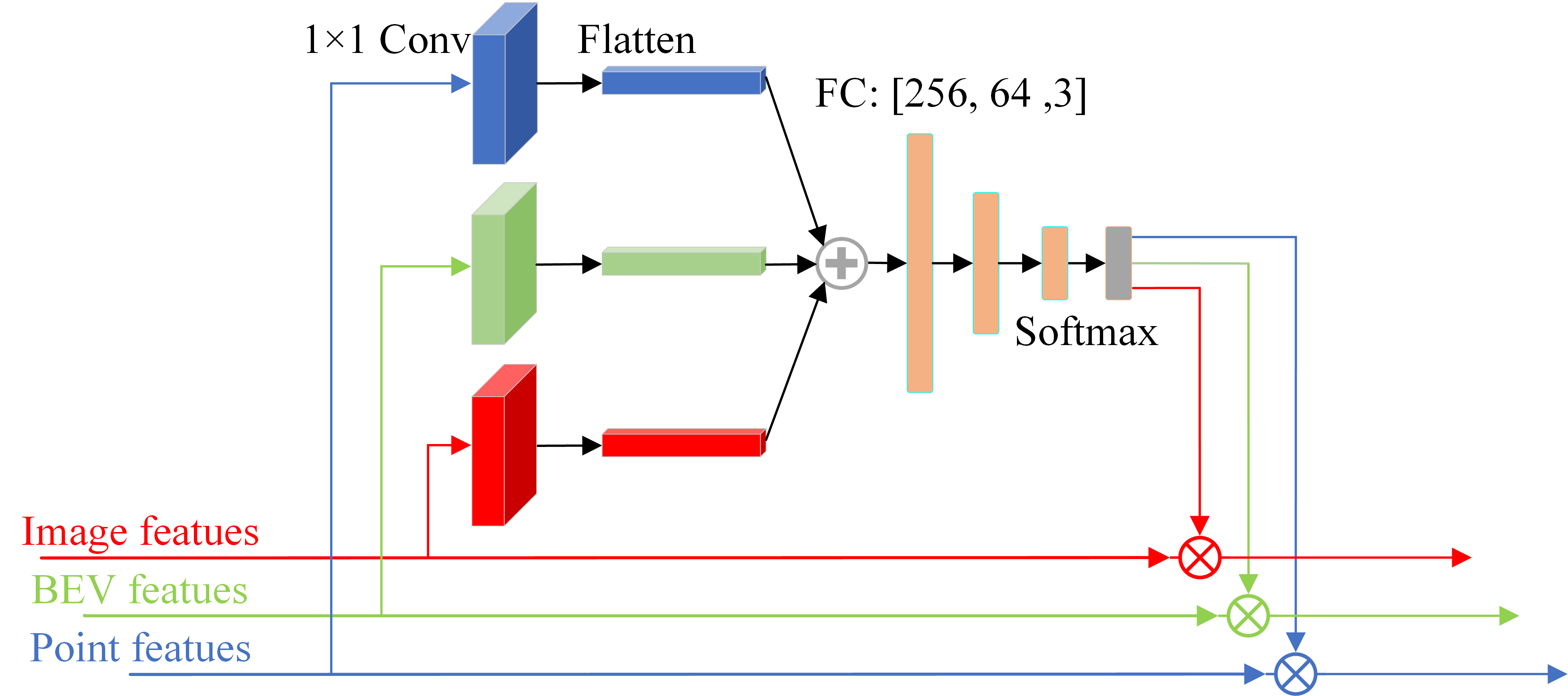}
\caption{Adaptive weighting module. Image features, BEV features and point features are shown in red, green and blue respectively.}
\label{AW}
\end{figure}
\paragraph{Unsupervised adaptive weighting module}
The goal of our Adaptive Weighting (AW) module is to balance three kinds of features mentioned above in an unsupervised manner. As we all know, the results of camera imaging and LiDAR scanning vary with the environment conditions like weather, as well as the conditions related to the objects such as pose, location and occlusion. Therefore, the importance of different features can be changeable depending on the context where the object lies. It is vital for a multi-sensor system to include a mechanism encouraging different data to compensate for each other.  
Here, we take image features, BEV features and point features as inputs to our AW module as shown in Figure \ref{AW}. 1 $\times$ 1 convolutional layers are used to reduce the dimension of features to a quarter of original ones. All the features are flatten to a one-dimensional vector and fused together by element-wise summation. A Multi-Layer Perceptron (MLP) with softmax function processes fused features and predicts three decimals between 0 and 1, which denote the weights of three kinds of features. Then the weights are multiplied with the input features to achieve balance between multiple signals for each proposal. 
\begin{figure*}
\centering
\includegraphics[scale=0.35]{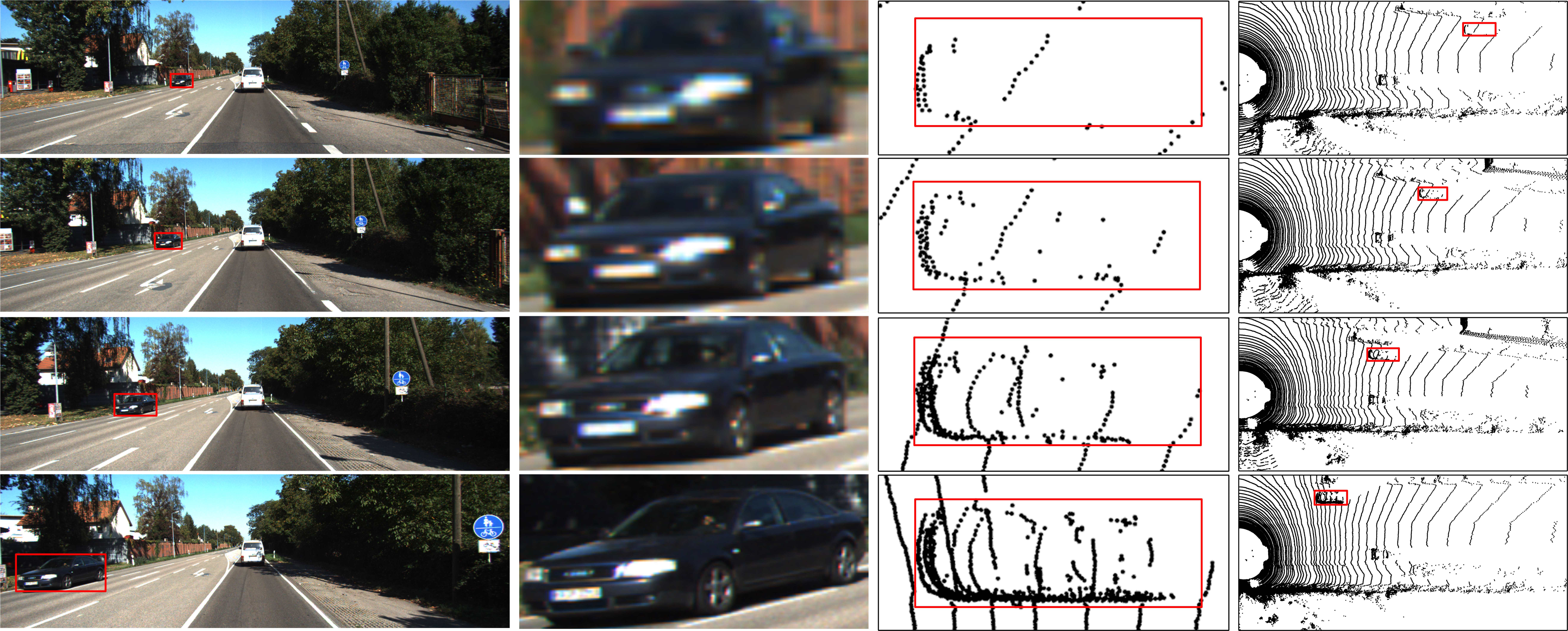}
\caption{Left part: Image and the cropped region of a moving car. Right part: LiDAR data and the cropped region of the car. As can be seen, the orientation shown in image region varies while remaining unchanged in LiDAR data. }
\label{AA}
\end{figure*}

\begin{figure}[t]
\centering
\includegraphics[scale=0.45]{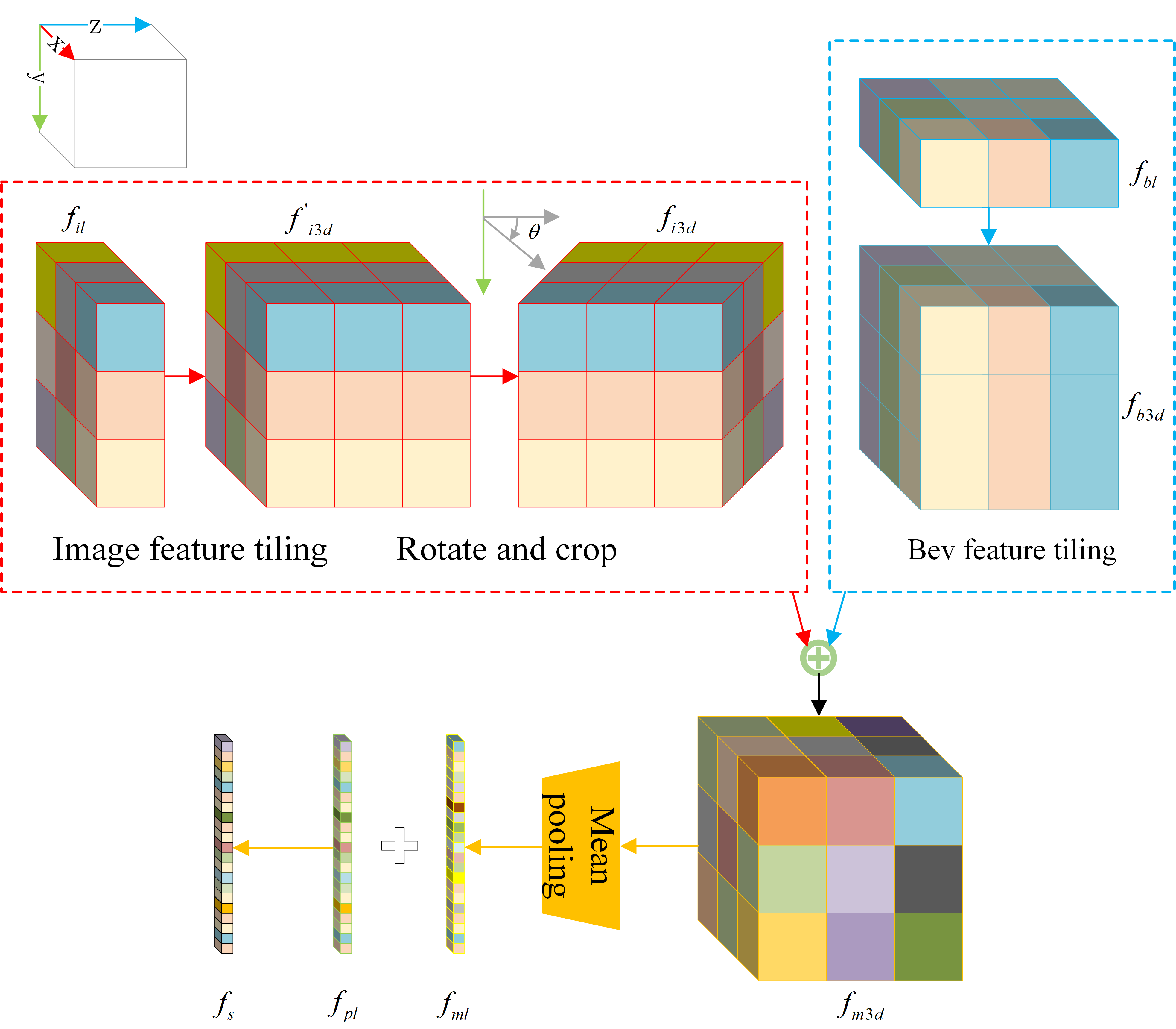}
\caption{Spatial fusion module. Parts in the red and blue box are the tiling process for image and BEV. Tiled image features are then rotated according to the azimuth between the proposal and the camera. To get axis-aligned features, we crop rotated image features and pad them into the same shape with BEV features. The cubes in the lower right are the merged features which contain information from image and BEV. Mean pooling operator reduces the dimension along x, y and z axis and sums features to output one-dimensional features that are then added with point features. }
\label{SF}
\end{figure}

\paragraph{Spatial fusion module}
The difference of local orientation shown in image and LiDAR data of the same object is illustrated in Figure \ref{AA}. Due to the change of azimuth between object and camera, local orientation shown in the image can be varied even when the object keeps the same direction in world coordinate system. In order to achieve an azimuth-aware fusion while considering the spatial information of the 3D object, we propose a novel Spatial Fusion (SF) module to project the 2D feature maps to a local 3D space to approximate the geometric structure of the 3D object and rotate image features according to the azimuth to achieve orientation-consistency with LiDAR features. The whole fusion process is shown in Figure \ref{SF}. In image tiling, we simply approximate the central projection of imaging process with orthographic projection based on the fact that the proposal usually occupies a minor part in the whole filed of the camera. 
Then $f_{i3d}$ and $f_{b3d}$ are summed up and we get a merged feature $f_{m3d}$ that incorporates the features from image and BEV. \textcolor{black}{To alleviate the computational burden caused by the additional dimension, we employ a mean-pooling layer to aggregate $f_{m3d}$ along x-axis, y-axis and z-axis.} The aggregated features are added together and flattened to a one-dimensional feature $f_{ml}$ with the same shape as $f_{pl}$ for later averaging fusion. The result of spatial fusion module is denoted as $f_{s}$, which includes information from image, BEV and point cloud features and has potential to promote the performance of the detection network. 
\paragraph{Final results and training objectives}
The fused features mentioned above are used to generate the final results. We divide the target into three sub-tasks, i.e., classification, corner prediction and angle prediction and define the objective for each task following \cite{ku2018joint}. Three fully-connected layers of size 2048 are used to further process the fused features which are shared by three sub-tasks. A Non-Maximum Suppression (NMS) with threshold of 0.01 is used to remove overlapping detection during inference stage.
Classification branch outputs the scores $Ref_{cls}$ for all classes including background. Cross-entropy loss is used for this task. For corner estimation branch, coordinates of four 2D corners and two heights are predicted to calculate 3D corners $Ref_{cor}$ as \cite{ku2018joint} and smooth L1 loss is applied here. Angle prediction branch gives the sine and cosine of the predicted angle denoted as $Ref_{ang}$ which are used to output the final angle prediction when combined with the estimation of four corners. Another smooth L1 loss is used for this branch. Loss of the refinement network can be formulated as:
\begin{align}
\begin{split}
Loss_{ref}=&\gamma_{c}*cross\_entropy(Ref_{cls},Ref_{cls}^{*})+\\ 
&\gamma_{r}*smoothL1(Ref_{cor},Ref_{cor}^{*})+\\
&\gamma_{a}*smoothL1(Ref_{ang},Ref_{ang}^{*}),
\end{split}
\end{align}%
where $Ref_{cls}^{*}$, $Ref_{cor}^{*}$ and $Ref_{ang}^{*}$ are ground-truth for three sub-tasks. We set the thresholds to 0.65 and 0.55 for positive and negative proposals according to their IoU with ground-truth box on BEV plane. To balance different tasks, $\gamma_{c}$, $\gamma_{r}$ and $\gamma_{r}$ are set to 1.0, 5.0 and 1.0 respectively. Only positive proposals involve the calculation of corner and angle loss. The details of each loss function are ignored because they are well elaborated in \cite{ku2018joint}. Our total loss can be denoted as follows:
\begin{align}
\begin{split}
Loss_{all}=Loss_{rpn}+Loss_{ref}+Loss_{gpen}.
\end{split}
\end{align}%

\section{Experiments}
\subsection{Performance evaluation for ground plane fitting}
To evaluate our ground plane fitting algorithm, we train the GPEN separately to get a fair comparison with other methods. KITTI dataset is split into training and validation set following \cite{chen2017multi}. To calculate pseudo ground plane parameters, we use the bounding boxes of all valid classes in each frame. Following Algorith \ref{alg1}, the ground plane vector $[a^*, b^*, c^*, d^*]$ is obtained for each frame. We train our GPEN wth a point set  $D=[x_{1},x_{2},x_{3},...,x_{k}]$ sampled from the preprocessed point cloud. In our experiments, $k$ is set to 521 and the model is trained for 30 epochs with an ADAM optimizer. Batch size is set to 32 which leads to about 2G GPU memory consumption. Learning rate is set to 0.001 initially and decays for every 10 epochs with a rate of 0.7.

\begin{figure}[t]
\centering
\includegraphics[scale=0.35]{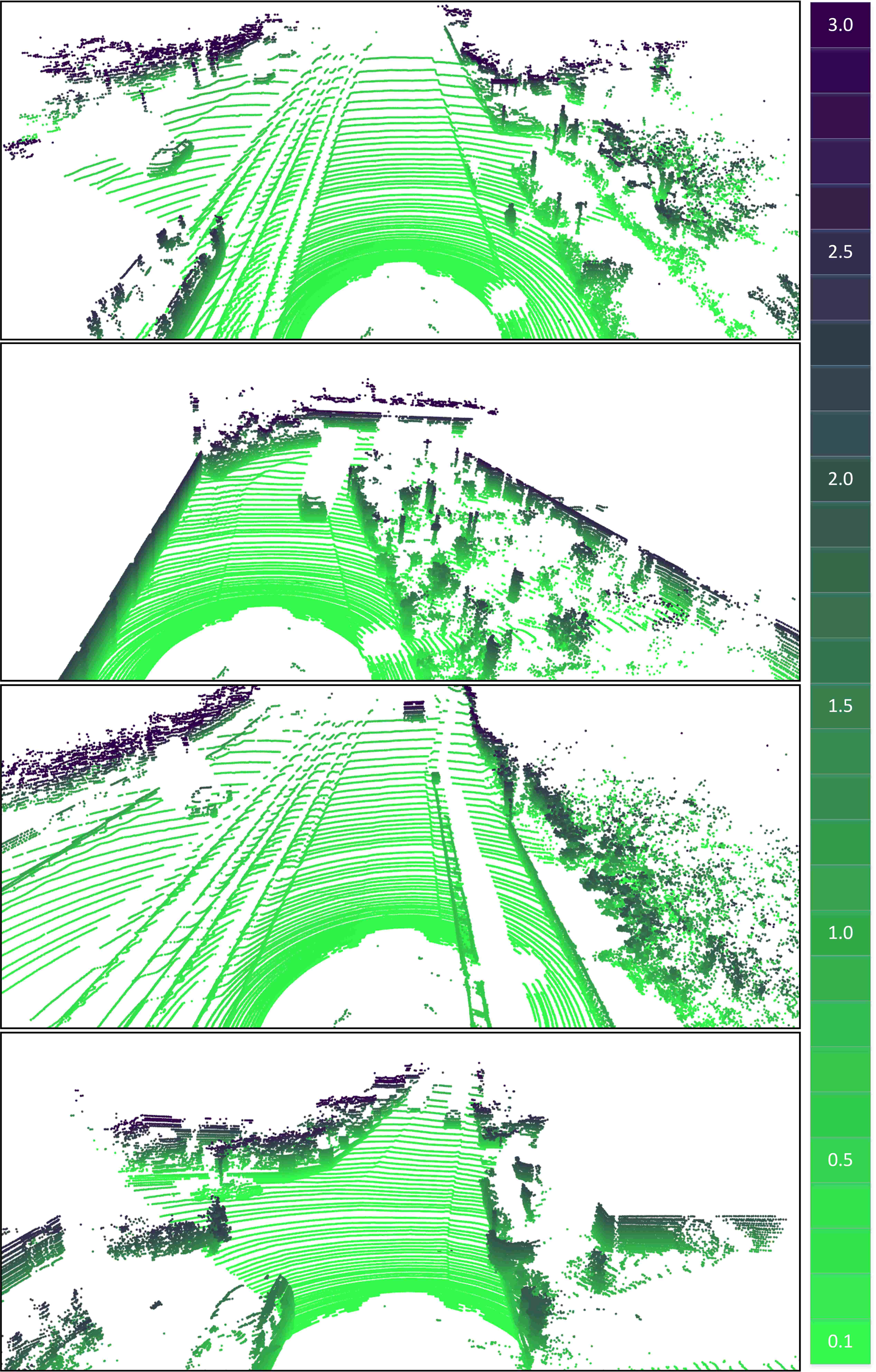}
\caption{Demos of our GPEN. We plot the location of each point and attach a color according to its distance to the predicted ground plane. The relationship of color and distance (in meters) is shown at the right side.}
\label{planes}
\end{figure}

Naive method means that we directly set the ground plane parameters to [0,-1,0,1.65]. It corresponds to the flat plane in camera coordinates whose normal vector is upward and distance to camera is 1.65 meters. For Least Square (LS) method, we minimize the square loss and get the corresponding parameters. For PCA method, we choose the eigenvector with the largest eigenvalue of the covariance matrix $XX^{T}$ as the normal vector $[a^{*},b^{*},c^{*}]$ of the plane and solve $d^{*}$ using the mean of all points,where $X$ is the matrix form of the point set for each frame. Actually, PCA method is equivalent with LS method here \cite{wang2001comparison}. We calculate the Root Mean Squard Error (RMSE) of angle (in degrees) and height (in meters). Angle error is defined as the deviation of angel between pseudo normal vector $[a^{*},b^{*},c^{*}]$ and predicted normal vectors $[a,b,c]$. The results are shown in Table \ref{Val_ground}. The comparison demonstrates the superiority of our method both on precision and speed. We can achieve 128 Frames Per Second (FPS) with a Titan X GPU (PASCAL) using Tensorflow. This makes it affordable when plugging our GPEN into other detection systems. Figure \ref{planes} shows the demos of the results of our GPEN.
\begin{table}
\centering
\begin{tabular}{cccc}
	\hline
	Method     & RMSE$_{ang}$ & RMSE$_{heig}$  & Speed(fps)\\
	\hline
	Naive      & 1.74   & 0.10       & --     \\
	LS         & 2.94   & 1.29      & 17.80     \\
	PCA \      & 2.95   & 1.29      & 18.10   \\
	RANSAC \cite{fischler1981random}     & 1.62   & 0.12     & 2.14  \\
	Ours       & \textbf{1.28}   & \textbf{0.10}     & \textbf{128.64}   \\   
	\hline
\end{tabular}
\caption{Comparison of our method with Naive method, Least Square (LS) method, Principal Components Analysis (PCA) method and RANSAC method on KITTI validation set for ground plane fitting.}
\label{Val_ground}
\end{table}
\begin{table}
\centering
\begin{tabular}{ccccc}
	\hline
	Method  & Type & Easy & Moderate & Hard\\
	\hline
	VoxelNet \cite{zhou2018voxelnet} &LiDAR & 81.97  & 65.46    & 62.85  \\
	SECOND \cite{yan2018second}    &LiDAR   & 87.43  & 76.48 & 69.10   \\
	\hline
	MV3D \cite{chen2017multi}   &Li-Cam     & 71.29  & 62.68 & 56.56  \\
	PointFusion \cite{xu2018pointfusion} &Li-Cam & 77.92  & 63.00    & 53.27  \\
	F-PointNet \cite{qi2018frustum}  &Li-Cam & 83.76  & 70.92 & 63.65  \\
	Cont Fuse \cite{liang2018deep} &Li-Cam  & 86.32  & 73.25 & 67.81  \\
	AVOD-FPN \cite{ku2018joint}   &Li-Cam  & 84.41  & 74.44 & 68.65  \\
	IPOD \cite{yang2018ipod}      &Li-Cam & 84.10   & 76.40  & 75.30   \\
	\textcolor{black}{MMF} \cite{liang2019multi} &Li-Cam      &  \textbf{87.90}  & \textbf{77.86}  & 75.57 \\
	\hline
	Ours (AW)   & Li-Cam  & 86.00 & 76.79 & 75.57 \\
	\textcolor{black}{Ours (AW+SF)} & Li-Cam & 86.77 & 76.84 & \textbf{75.92} \\   
	\hline
\end{tabular}
\caption{Comparison of our method with other state-of-the-art approaches on the car class of KITTI validation set for 3D detection. Type ``LiDAR" denotes methods that use LiDAR data only and ``Li-Cam" marks methods that adopt LiDAR-camera setup.  }
\label{Val_3D}
\end{table}
\begin{table}
\centering
\begin{tabular}{ccccc}
	\hline
	Method  & Type & Easy & Moderate & Hard\\
	\hline
	VoxelNet \cite{zhou2018voxelnet}  & LiDAR & 89.60  & 84.81    & 78.57  \\
	SECOND \cite{yan2018second}   &LiDAR    & 89.96 & 87.07 & 79.66 \\
	\hline
	MV3D \cite{chen2017multi}   &Li-Cam     & 86.55 & 78.10  & 76.67 \\
	PointFusion \cite{xu2018pointfusion} &Li-Cam  & 87.45 & 76.13 & 65.32 \\
	F-PointNet \cite{qi2018frustum} &Li-Cam & 88.16 & 84.02 & 76.44 \\
	Cont Fuse \cite{liang2018deep} &Li-Cam   & 95.44 & 87.34 & 82.43 \\
	AVOD-FPN \cite{ku2018joint}   &Li-Cam   & -     & -     & -     \\
	IPOD \cite{yang2018ipod}    &Li-Cam     & 88.30  & 86.40  & 84.60  \\
	\textcolor{black}{MMF} \cite{liang2019multi} &Li-Cam      &  \textbf{96.66}  & \textbf{88.25}  & 79.6  \\
	\hline
	Ours (AW)   &Li-Cam & 89.66 & 86.95 & 79.77 \\
	\textcolor{black}{Ours (AW+SF)} &Li-Cam & 89.95 & 87.70 & \textbf{86.95} \\   
	\hline
\end{tabular}
\caption{Comparison of our method with other state-of-the-art approaches on the car class of KITTI validation set for BEV detection (localization). Type ``LiDAR" denotes methods that use LiDAR data only and ``Li-Cam" marks methods that adopt LiDAR-camera setup.}
\label{Val_BEV}
\end{table}
\begin{table}
\centering
\begin{tabular}{cccc}
	\hline
	Method   & Easy & Moderate & Hard\\
	\hline
	AVOD-FPN \cite{ku2018joint}      & 84.19     & 74.11     & 68.28     \\
	Ours (AW)    & 85.89 & 76.40 & 74.88 \\
	\textcolor{black}{Ours (AW+SF)} & \textbf{86.75} & \textbf{76.53} & \textbf{74.90} \\   
	\hline
\end{tabular}
\caption{Comparison of our method with AVOD on the car class of KITTI validation set for AHS.}
\label{Val_AHS}
\end{table}
\subsection{Performance evaluation for 3D detection}
We train our network on the KITTI dataset and focus on the car class like \cite{chen2017multi,liang2018deep,shin2018roarnet} because it has the most samples among all classes of the dataset. The dataset is split into training and validation set following \cite{chen2017multi}. We use the same RPN module as \cite{ku2018joint}. The number of proposals is set to 1024 during training and 300 for validation. For feature extraction, point features are resized to [128, 1568] and both image and BEV featuresare cropped and resized into $[7,7,32]$ for each proposal. The batch size is set to 1 considering the memory usage, and learning rate is set to 0.0001 that decays exponentially with a factor of 0.8 for every 30k iterations. We train our network for 120k iterations with an ADAM optimizer. \textcolor{black}{The inference speed of our model on KITTI dataset is 0.12 second per frame on our workstation with an Intel Xeon CPU (E5-2650@2.2GHz) and a NVIDIA Titan X GPU (PASCAL).}

\begin{figure}
\centering
\includegraphics[width=3.1in]{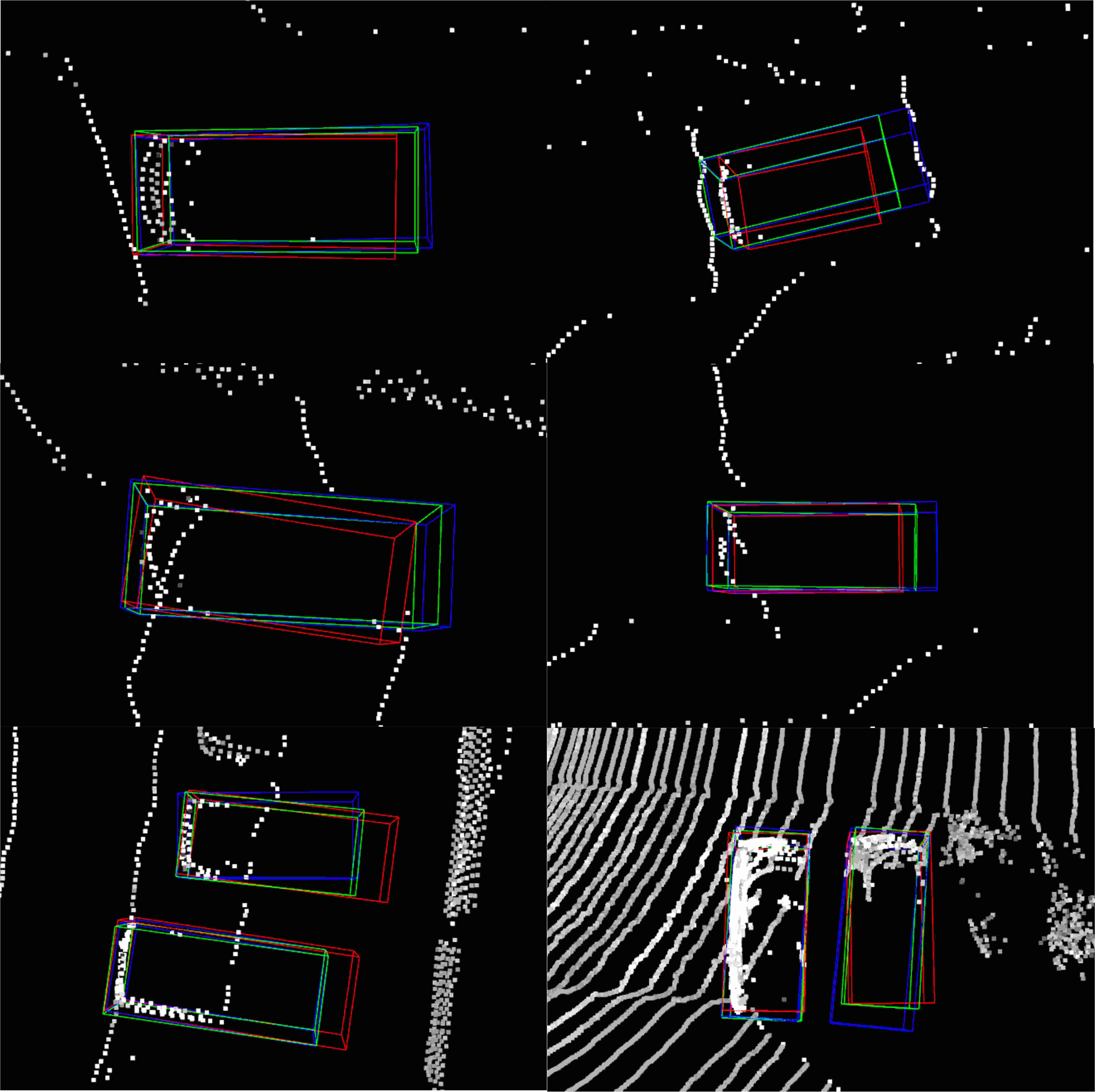}
\caption{Visualization of the baseline and our detection results. Boxes in red denote the ground truth. Boxes in blue are the results of our baseline which only uses image and BEV features. Boxes in green are the results of our method that adds point features.}
\label{demo2}
\end{figure}

\begin{figure*}
\centering
\includegraphics[width=6.0in]{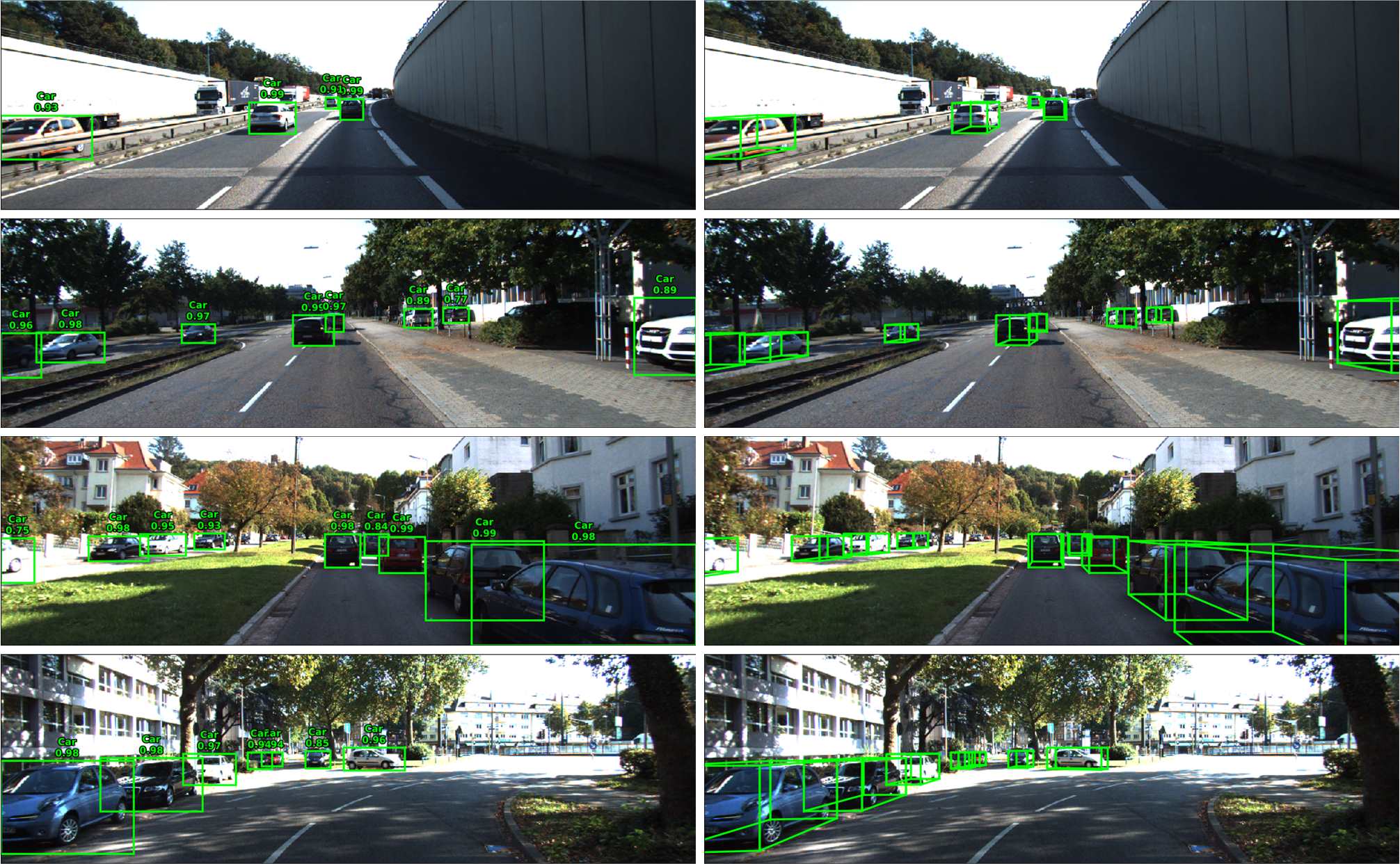}
\caption{Visualization of our detection results. 2D detection results are shown in the left side and 3D detection results are shown in the right side.}
\label{demo}
\end{figure*}

For comparison with other state-of-the-art approaches \cite{zhou2018voxelnet,yan2018second,chen2017multi,ku2018joint,liang2018deep,xu2018pointfusion,qi2018frustum,yang2018ipod}, we evaluate our method on the KITTI validation set using Average Heading Similarity (AHS) and Average Precision (AP) at 0.7 3D IoU. Table \ref{Val_3D} and Table \ref{Val_BEV} show the results of AP on 3D detection and BEV detection respectively. For 3D detection under moderate mode, we are 2.40\% higher than AVOD \cite{ku2018joint} which uses image and BEV features and 5.92\% higher than F-PointNet \cite{qi2018frustum} which mainly relies on point features. It demonstrates that aggregating richer local features from image, BEV and point cloud can promote the performance of 3D detection. Although PointFusion \cite{xu2018pointfusion} and Cont Fuse \cite{liang2018deep} also design particular modules for the fusion of LiDAR and image data, their performance are 13.84\% and 3.59\% lower than ours respectively under moderate mode, which shows the advantages of our proposed fusion method. For BEV detection, we also get ahead on both moderate and hard mode. Recent work MMF \cite{liang2019multi} demonstrates a strong approach that incorporates both image and LiDAR features. The main contributions of MMF \cite{liang2019multi} include the design of supplementary tasks to extract better features and the fusion of multimodal features on both point-level and ROI-level. By contrast, we focus more on the generation of diverse features from LiDAR data \textcolor{black}{to maintain more 3D informaiton} and propose an attention-based fusion mechanism when aggregating image and LiDAR features. MMF uses image and BEV features to make the final prediction while we also extract point features from original point cloud to maintain more spatial information. Compared with the addition or concatenation used by MMF to fuse different features, our adaptive weighting network can dynamically adjust the importance of different features and  improve the representation ability of the fusion module. MMF gets better performance than ours on easy and moderate modes while our approach shows advantage in hard mode. Considering the complexity of traffic dataset like KITTI, hard examples usually lack supportive evidence to be precisely detected. \textcolor{black}{Current voxelization method and downsampled BEV-based features can further aggravate the situation of hard cases. Voxelzation operation merges the points inside the same voxel into a single vector which inevitably loss part of the 3D information. And BEV-based network generally makes the final detection on the low-resolution feature maps which are downsampled by 4 or 8 times compared with the original input. Both of them can degrade the localization ability of the 3D detection network especially for hard cases. Our point-based branch well makes up for the information loss caused by voxelization and feature downsampling. We extract point features directly from the original LiDAR data to maintain the full 3D information and combine them with BEV and image features. This helps us to make preciser localization as shown in Fig \ref{demo2}.} We think that is the advantage of our method and the reason why we get better results than MMF for hard cases. The comparison of the heading estimation performance with AVOD is shown in Table \ref{Val_AHS}. The model with AW and SF modules beats that with only AW for easy and moderate modes, which shows that the introduction of orientation consistency between different features helps to achieve better angle estimation. Figure \ref{demo} visualizes our detection results on some samples in KITTI dataset.
\begin{table}[t]
\centering
\begin{tabular}{cccccc}
	\hline
	\textcolor{black}{PL} &PointNet  & AW & SF & $3D_{M}$ & $BEV_{M}$\\
	\hline
	$\times$& $\surd$  & $\surd$  & $\surd$  & 65.10 & 78.28 \\
	$\surd$ & $\times$  & $\times$  & $\times$ & 72.48 & 85.70 \\
	$\surd$ & $\surd$  & $\times$  & $\times$ & 73.18 & 86.15 \\
	$\surd$ & $\surd$  & $\times$  & $\surd$ & 73.72 & 86.49\\
	$\surd$ & $\surd$  & $\surd$  & $\times$ & 76.79 & 86.95\\
	$\surd$ & $\surd$  & $\surd$  & $\surd$ & \textbf{76.84} & \textbf{87.70} \\
	\hline
\end{tabular}
\caption{Results on the car class of KITTI validation set for 3D and BEV detection when using different setups of the proposed modules \textcolor{black}{(``PL" means plane loss, ``AW" means adaptive weighting module and ``SF" means spatial fusion module)}.}
\label{Abl_module}
\end{table}
\begin{table}[t]
\centering
\begin{tabular}{ccccc}
	\hline
	image  & BEV & point & $3D_{M}$ & $BEV_{M}$\\
	\hline
	$\surd$  & $\surd$  & $\times$ & 72.48 & 85.70 \\
	$\surd$  & $\times$  & $\surd$ & 73.80 & 86.07 \\
	$\times$  & $\surd$  & $\surd$ & 53.53 & 67.14\\
	$\surd$  & $\surd$  & $\surd$ & \textbf{76.79} & \textbf{86.95}\\
	\hline
\end{tabular}
\caption{Results on the car class of KITTI validation set for 3D and BEV detection when using different combinations of features in the refinement network.}
\label{Abl_feature}
\end{table}
\begin{table}[t]
\centering
\begin{tabular}{cccc}
	\hline
	Reduction method & $3D_{M}$ & $BEV_{M}$ & $AHS_{M}$\\
	\hline
	Max-Pool (yz)   & 76.00 & 86.74 & 75.95\\
	Mean-Pool (yz)   & 76.74 & 87.49 & 76.48\\
	Mean-Pool (xyz)   & \textbf{76.84} & \textbf{87.70} & \textbf{76.53}\\
	\hline
\end{tabular}
\caption{Results on the car class of KITTI validation set for 3D detection, BEV detection and AHS with different reduction methods in our SF module.}
\label{Abl_reduction}
\end{table}
\subsection{Ablation study}
\begin{figure}
\centering
\includegraphics[width=3.2in]{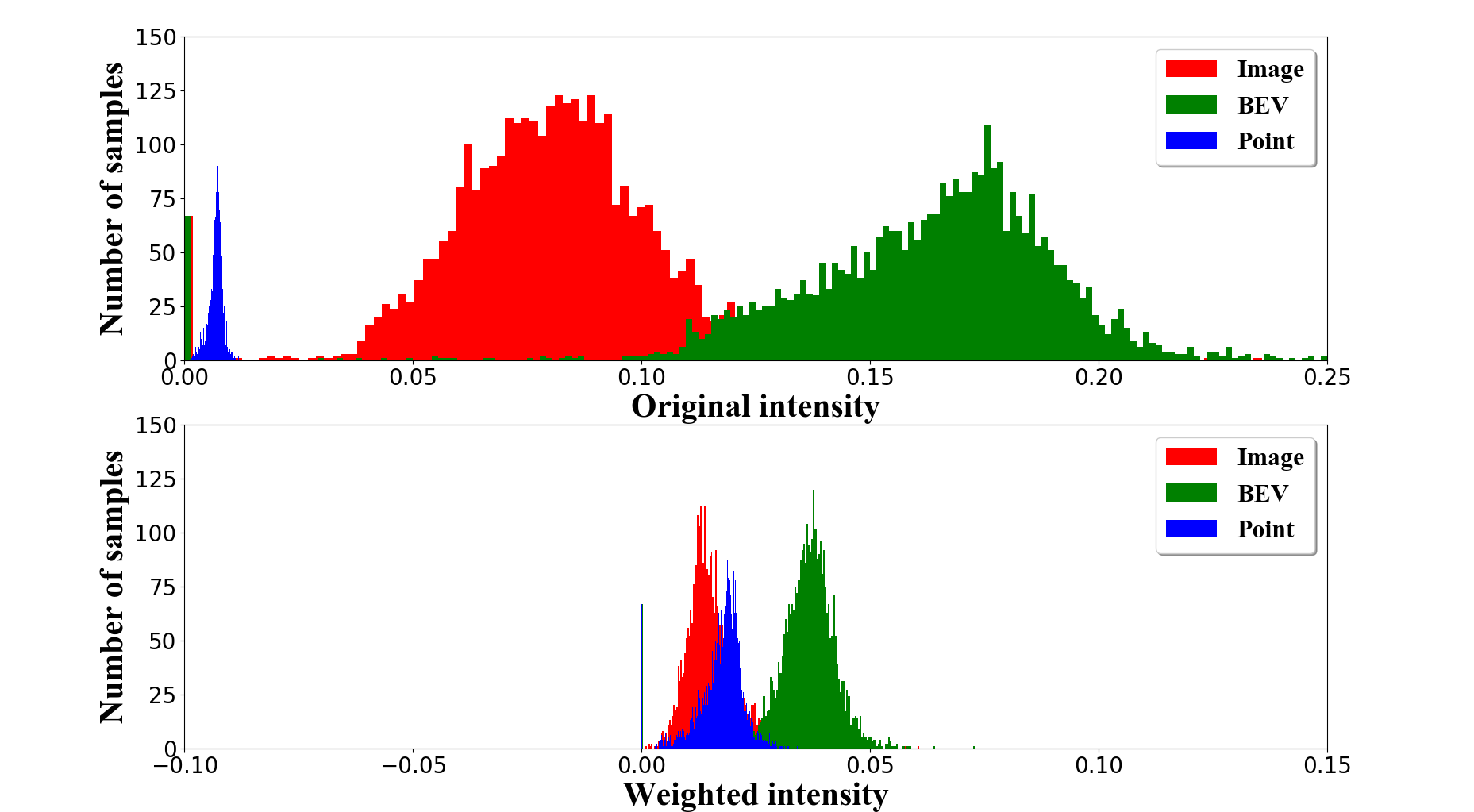}
\caption{\textcolor{black}{Distributions of the intensity of different features on the car class of KITTI validation set. The upper subfigure shows the intensity of original features and the lower one shows that of features after our adaptive weighting module.}}
\label{dist_feature}
\end{figure}

In this part, we analyze the effectiveness of our proposed modules and the function of different features used in our model, i.e. image features, BEV features and point features. For clarity, we use three kinds of features for every experiment in Table \ref{Abl_module}. Due to the fact that SF module necessarily feeds on all these features, we delete the SF module for experiments in Table \ref{Abl_feature} and evaluate the function of different features in the refinement network.  All the models are trained on KITTI training set and evaluated on the validation set for 3D and BEV detection under moderate mode.
Table \ref{Abl_module} shows the results when we use different combinations of the proposed modules. \textcolor{black}{To illustrate the necessity of ground plane estimation, we remove the plane loss in our network and use naive plane parameters ([0,-1,0,1.65]) to generate anchors. In this case, PointNet doesn't predict ground plane parameters and is used only for the point feature extraction. Without plane loss, the detection performance lags 11.74\% and 9.42\% behind the full model, thereby verifying the necessity to make precise ground plane estimation. It is important especially for large scenes like traffic scene where a small angle deviation of the ground plane can lead to significant gaps between objects and their corresponding anchors.}  The experiments with different combination of modules demonstrates our intention to improve the performance of 3D object detector by making full use of the LiDAR data and fusing it with image features effectively. In detail, the introducing of PointNet to extract point features increases the  performance for both 3D and BEV detection. Although the BEV features have a good representation of the LiDAR data \cite{ku2018joint}, it is better to keep the original point cloud. This conclusion is further verified in Table \ref{Abl_feature}, where 3D and BEV detection AP increase 1.32\% and 0.37\% respectively when we replace BEV features with point features. It shows that features extracted from the original 3D point cloud are more suitable for 3D refinement which may alleviate the loss of useful information when processing LiDAR data thanks to recent point-based feature extractors \cite{qi2017pointnet,qi2017pointnet++}. Secondly, AW and SF modules compose a more effective fusion style. 3D detection AP grows by 3.61\% when the AW module is added. This illustrates the importance to adjust the strength of different features especially when we face heterogeneous data. For different features of positive proposals on KITTI validation set, we plot distributions of their intensity as shown in Figure \ref{dist_feature}. Intensity of features is calculated by averaging their values. It is clear to see the huge difference on intensity between original features. Stronger ones like BEV features can easily cover those with a weak strength like point features and dominate the training process thus leading to a vulnerable model. Our AW module changes this situation by dynamically adjusting the intensity and achieves a balance between different features. From Table \ref{Abl_module}, we can also see that AW module is not only beneficial but also necessary for the following SF module because only adding SF to the PointNet-deployed network does not give a significant increase. It illustrates again the importance of balancing multimodal data. 

Results in Table \ref{Abl_feature} show a large drop of AP after we remove image features, which can probably be interpreted by the important role of image in classification task of our framework since we witness massive missed detection without image features. Table \ref{Abl_feature} also shows that although redundancy lies between BEV and point features, retaining them at the same stage helps to promote the performance of 3D detection. 

\textcolor{black}{In Table \ref{Abl_reduction}, we show the performance of different pooling methods to reduce 3D features ($f_{m3d}$ in Figure \ref{SF}) into 2D ones in SF module. As shown in Figure \ref{SF}, mean pooling achieves better results than max pooling when we pool 3D features along the y axis and z axis. One advantage of mean pooling is that it keeps the original features before tiling. Taking the pooling along y axis as an example, the generated feature $f_{pooly}$ is:
\begin{align}
\begin{split}
f_{pooly}&=MeanPool_y(f_{m3d})\\&=MeanPool_y(f_{i3d}+f_{b3d})\\&=MeanPool_y(f_{i3d})+MeanPool_y(f_{b3d})\\&=MeanPool_y(f_{i3d})+f_{bl},
\end{split}
\end{align}%
where $MeanPool_y$ is the mean pooling along y axis. When the pooling results along x axis is added, we get better results. It shows the new perspective (x axis) benefits the perception of 3D objects.
}
\section{Conclusion}
In this paper, we propose a 3D detection network that aggregates rich local features from image, BEV maps and point cloud. We design a ground plane fitting network to estimate the ground parameters and produce point cloud features.
Image and BEV maps are processed in RPN by 2D CNNs to generate image-like features. With such a design, we utilize mature 2D CNNs and point-based 3D extractors to explore the potential of LiDAR data for 3D object detection. Besides, our adaptive fusion network provides an effective way to fuse features from multimodal data. The adaptive weighting module adjusts the strength of each signal and chooses information for later operation, and the spatial fusion module incorporates the azimuth and geometry information into the mergence of multiple features. Experimental results on KITTI dataset illustrate the validity of our method. \textcolor{black}{In the future, we plan to reduce the reliance on the plane assumption and extend our approach to fit curved ground and build a more robust network.}

\section{Acknowledgements}
This work was supported partly by The Intel Collaborative Research Institute for Intelligent and Automated Connected Vehicles (ICRI-IACV), and National Natural Science Foundation of China (No. 61533019, U1811463).








\bibliographystyle{elsarticle-num} 
\bibliography{elsarticle-template-num}

\end{document}